\documentclass[]{interact}

\usepackage{url}
\usepackage{setspace}
\usepackage{tikz}
\usepackage{multirow, multicol}                   
\usepackage{mathrsfs,amsmath,amsfonts,amssymb}
\usepackage{algorithm, algorithmic}
\usepackage{fancyhdr}
\usepackage{color}
\usepackage{subfigure}
\usepackage{lmodern}
\usepackage[T1]{fontenc}
\usepackage{MnSymbol}
\usepackage{hyperref}
\hypersetup{
    colorlinks=true,
    linkcolor=blue,
    filecolor=magenta,
    urlcolor=cyan,
}
\usepackage{epstopdf}

\usepackage[numbers,sort&compress]{natbib}
\theoremstyle{plain}
\newtheorem{theorem}{Theorem}[section]
\newtheorem{proposition}[theorem]{Proposition}
\usepackage{amsmath}
\usepackage{amsthm}
\usepackage{amssymb}

\theoremstyle{definition}
\newtheorem{definition}[theorem]{Definition}
\newtheorem{example}[theorem]{Example}
\usepackage{hyperref}
\usepackage{algorithm, algorithmic}
\usepackage{multirow}                       
\usepackage{mathrsfs,amsmath,amsfonts,amssymb}
\usepackage{color}
\usepackage{anyfontsize}
\usepackage{MnSymbol}
\theoremstyle{remark}
\newtheorem{remark}{Remark}
\usepackage{caption}

\def\hat{\widehat}
\def\tilde{\widetilde}

\def\epsilon{\varepsilon}

\def\*{$\!\!^{^{^{\displaystyle *}}}$}

\newcommand{\vesub}[2]{\mbox{{\boldmath ${#1}$}$_{#2}$}}

\def\Rmn#1{\expandafter\uppercase\expandafter{\romannumeral #1}}

\begin{document}

\title{Intrinsic Gaussian Process Regression Modeling for Manifold-valued Response Variable}
\author{\name{Zhanfeng Wang\textsuperscript{a} and Xinyu Li\textsuperscript{a} and Hao Ding\textsuperscript{a} and Jian Qing Shi\textsuperscript{b} 
\thanks{*Corresponding authors: Jian Qing Shi, E-mail: shijq@sustech.edu.cn.}}
\affil{\textsuperscript{a} International Institute of Finance, School of Management, University of Science and Technology of China, Hefei, 230026, Anhui, China.\\
\textsuperscript{b} Department of Statistics and Data Science, College of Science, Southern University of Science and Technology, Shenzhen, 518055, Guangzhou, China.}}

\maketitle

\begin{abstract}
Extrinsic Gaussian process regression methods, such as wrapped Gaussian process, have been developed to analyze manifold data. However, there is a lack of intrinsic Gaussian process methods for studying complex data with manifold-valued response variables. In this paper, we first apply the parallel transport operator on Riemannian manifold to propose an intrinsic covariance structure that addresses a critical aspect of constructing a well-defined Gaussian process regression model.
We then propose a novel intrinsic Gaussian process regression model for manifold-valued data, which can be applied to data situated not only on Euclidean submanifolds but also on manifolds without a natural ambient space. We establish the asymptotic properties of the proposed models, including information consistency and posterior consistency, and we also show that the posterior distribution of the regression function is invariant to the choice of orthonormal frames for the coordinate representations of the covariance function. Numerical studies, including simulation and real examples, indicate that the proposed methods work well.
\end{abstract}
\begin{keywords}
Gaussian process regression, Riemannian manifold, Intrinsic analysis, Parallel transport.
\end{keywords}

\section{Introduction}
Regression analysis is a fundamental statistical technique that is used to uncover the relationship between independent and dependent variables. Most well-known regression models are developed under the assumption that these variables reside within Euclidean spaces. However, with recent advances in science and technology, there is a growing collection of data showing a non-linear manifold structure. Consequently, statistical models of manifold-valued data are becoming increasingly important in various scientific domains, including computer vision, neuroscience, and shape data analysis \citep{bronstein2017geometric}. Since traditional regression models based on Euclidean space are inadequate for capturing the intrinsic geometric properties of manifold-valued data, this paper aims to study Gaussian process regression models using a novel intrinsic covariance function applied to complex data with manifold-valued response variable. 

In the regression framework, there are generally two types of data with multiple values: one for covariates and the other for the response variable. Current research focuses mainly on the case involving covariates \citep{davis2010population,article,banerjee2015nonlinear,lin2017extrinsic,pelletier2006non,lin2021functional,dai2018principal,lin2019intrinsic,azangulov2022stationary,banerjee2016nonlinear,borovitskiy2020matern}. For example, \cite{lin2019intrinsic} proposed a linear regression model with functional predictors on manifolds. \cite{azangulov2022stationary} extended the stationary Gaussian process model to compact Lie groups and non-compact symmetric spaces.  \cite{banerjee2016nonlinear} introduced a kernel-based nonlinear regression model for independent variables valued at manifolds. \cite{borovitskiy2020matern} utilized the spectral theory of the Laplace-Beltrami operator to compute Matérn kernels on Riemannian manifolds and developed a nonparametric regression model for predictors valued at the manifold. Although these studies investigate regression models with predictors with multiple values, the response variables typically reside in Euclidean space. However, there are scenarios where the responses are directly situated on Riemannian manifolds. For example, many studies explore how brain shape changes with age, demographic factors, IQ, and other variables \citep{cornea2017regression}. It is essential to consider the underlying geometry of the manifold for an accurate inference. Ignoring the geometry of the data can potentially lead to highly misleading predictions and conclusions. To the best of our knowledge, this is the first investigation of an intrinsic Gaussian process designed for manifold-valued data. 
We aim to develop new statistical methods to study the intrinsic covariance structure of a manifold-valued variable.

Since the result of sum and product are usually not on the originated manifold, it is very difficult to define a regression model on a specific manifold. Research in this area is relatively rare \citep{thomas2013geodesic,nava2020geodesic,hinkle2012polynomial,petersen2019frechet,bhattacharjee2023single,lin2023additive}. For example, geodesic regression \citep{thomas2013geodesic} is a direct generalization of linear regression, designed to find the relationship between a predictor variable with real value and a response variable with manifold value via a geodesic curve on the manifold. \cite{nava2020geodesic} provided an analytic approach to geodesic regression in Kendall’s shape space. In addition to geodesic regression, other methodologies have also been developed to analyze manifold-valued response variables. For example, \cite{hinkle2012polynomial} extended polynomial regression to Riemannian manifolds by introducing a class of curves generalizing geodesics, which are analogous to polynomial curves in Euclidean space.  \cite{petersen2019frechet} proposed generalized versions of both global least squares regression and local weighted least squares smoothing methods when responses are complex random objects in a metric space and predictors reside in Euclidean space. \cite{bhattacharjee2023single} proposed a novel single index regression model that involved responses to random objects of metric space values coupled with multivariate Euclidean predictors. \cite{lin2023additive} investigated an additive regression model for symmetric positive-definite matrix-valued responses with multiple scalar predictors. Nevertheless, the aforementioned works focus mainly on the models with certain parametric structures (e.g., a linear or a generalized linear model ). In practice, we are also interested in modeling nonlinear relationship nonparametrically.  A way is to estimate a distribution for potential regression functions. The intrinsic regression model proposed in this paper is designed to address this problem.

Gaussian process regression (GPR) is a powerful tool to model nonlinear and non-parametric relationship between response variable and covariates \citep{shi2011gaussian}. The idea of GPR has also been applied to model data on the manifold. For example, \cite{pigoli2016kriging} introduced a kriging method for the manifold-valued response variable which is based on the exponential map. Their method specifically uses multivariate geodesic regression to form a reference coordinate system used to compute residuals of the manifold-valued data points. Regular GP regression is then applied to model the residuals, and the results are mapped back onto the manifold.  However, the kriging method relies heavily on localizing the data to a single tangent space and cannot offer an intrinsic probabilistic interpretation. 
\cite{mallasto2018wrapped} and \cite{liu2024wrapped} proposed a wrapped Gaussian process regression (WGPR), which linearizes the Riemannian manifold via an exponential map to a tangent space, providing a nonparametric regression framework with probabilistic interpretation. WGPR relaxes some constraints of the kriging method by utilizing wrapped Gaussian distribution (WGD) designs and prior information such as geodesic submanifold. 
However, this approach may not be suitable for general settings. 
First, it is not suitable to fit manifolds which are not a Euclidean submanifold or do not have a natural isometric embedding into a Euclidean space. For example, the Riemannian manifold of $p \times p$ symmetric positive definite (SPD) matrices endowed with the affine-invariant metric is not compatible with the $p(p+1)/2$-dimensional Euclidean metric. The WGPR cannot be applied to those types of manifolds. Second, although an ambient space can provide a common stage for tangent vectors at different points, operation computation, such as additive and inner product, on tangent vectors from the ambient perspective may potentially violate the intrinsic geometry of the manifold.  In addition, since manifolds might be embedded into multiple ambient spaces, the interpretation of statistical results heavily relies on the choice of ambient space and could be misleading if not chosen appropriately. There are other works considered the distribution for potential regression functions through Kalman filtering models \citep{hong2017regression} and stochastic development \citep{kuhnel2017stochastic}. 
As an alternative, GPR offers a more flexible way to model regression functions in a nonlinear and nonparametric manner.
By building an appropriate kernel function for manifold-valued variables, GPR can effectively incorporate the information from either Euclidean or manifold-valued predictors to model manifold-valued responses. Nevertheless, to our knowledge, no method exists for studying the intrinsic GPR model for manifold-valued data.
We aim to fill this gap by developing an intrinsic method for covariance kernel function in Riemannian manifold spaces. 

In this paper, we propose a new Gaussian process regression model for manifold-valued responses that rigorously preserve the intrinsic geometry of manifolds, denoted by the intrinsic Gaussian process regression model (iGPR). Our methodological and theoretical contributions include the following:
(1) We propose a new intrinsic covariance structure based on a parallel transport operator which can be applied to various important Riemannian manifolds that are not naturally Euclidean submanifolds but are frequently encountered in statistical analysis and machine learning, such as the aforementioned SPD manifolds. The innovation use of the parallel transport operator 
allows the covariance structure to preserve the intrinsic geometry information of the manifold. (2) We apply the developed covariance function to propose an intrinsic Gaussian process regression model for manifold-valued responses. We also demonstrate that the estimation results are invariant to the choice of a common tangle space and coordinate frames, although the representation of the covariance function needs a common tangle space and coordinate frames. (3) We apply a convenient coregionalization model to develop an intrinsic method to estimate the covariance function. We further develop an efficient computational method for iGPR to obtain the predictive distribution of the regression function.
(4) Under some mild conditions, we establish the asymptotic properties of the proposed models, including information consistency and posterior consistency. Simulation studies and real data analysis highlight the effective performance of the proposed method.

The remainder of this paper is organized as follows. In section \ref{s1}, we review key concepts of Riemannian manifolds are reviewed, and propose an intrinsic Gaussian process regression model for manifold-valued response variables along with the asymptotic properties of the proposed model. Sections \ref{s2} and \ref{section:4} report the result of simulation experiments and real data analysis. Finally, Section \ref{s3} offers concluding remarks. All proofs and more numerical studies are provided in the Supplementary Materials.

\section{Methodology}\label{s1}
\subsection{Background and notations}
A $D$-dimensional Riemannian manifold $\mathcal{M}$ is a $m$-dimensional differentiable and smooth manifold equipped with a Riemannian metric which defines an inner product $\langle\cdot, \cdot\rangle_p$ on the tangent space $T_p \mathcal{M}$ at each point $p \in \mathcal{M}$.  Denoted the associated norm by $\|v\|_p=\sqrt{\langle v, v\rangle_p}$ for $v \in T_p \mathcal{M}$. The metric, which smoothly varies with $p$, induces a distance function $d_{\mathcal{M}}$ on $\mathcal{M}$ and turns the manifold into a metric space. If such a metric space is complete, then $\mathcal{M}$ is called a complete Riemannian manifold. A geodesic in a Riemannian manifold is a constant-speed curve, where every sufficiently small segment is the shortest path connecting the endpoints of the segment. In addition, a Riemannian manifold $\mathcal{M}$ is connected if it is not the union of two disjoint nonempty open sets.

Assume that the Riemannian manifold $\mathcal{M}$ is complete and connected, the Riemannian metric also induces the Riemannian exponential map $\operatorname{Exp}(p,\cdot)$ for each $p \in \mathcal{M}$, which maps tangent vectors at $p$ onto the manifold $\mathcal{M}$. The inverse of $\operatorname{Exp}(p,\cdot)$, when it exists, is called the Riemannian logarithmic map at $p$ and denoted by $\operatorname{Log}(p,\cdot)$. Moreover, if the Riemannian exponential map is surjective at every point $p \in \mathcal{M}$, it implies that $\mathcal{M}$ is geodesically complete based on the Hopf-Rinow theorem \citep{ekeland1978hopf}. However, given two points on $\mathcal{M}$, the geodesic between the points may not be unique, even if $\mathcal{M}$ is geodesically complete. Consider the ball of the largest radius around the origin in $T_p \mathcal{M}$, on which $\operatorname{Exp}(p, v)$ is defined, and let $V(p) \subset \mathcal{M}$ denote the image of $\operatorname{Exp}(p, \cdot)$ on this ball. Then the exponential map has an inverse on $V(p)$, called the Riemannian logarithm map at $p, \operatorname{Log}(p, \cdot): V(p) \rightarrow T_p \mathcal{M}$. The injectivity radius at $p \in \mathcal{M}$, denoted $\operatorname{inj}_p$, is defined as the radius of the largest ball at the origin of $T_p \mathcal{M}$ in which $\operatorname{Exp}(p, \cdot)$ is a diffeomorphism. It follows that the geodesic ball at $p$ of radius $\operatorname{inj}_p$ is contained within $V(p)$. 
More details can be found in \cite{pennec2006intrinsic}. In this paper, data are assumed from a Riemannian manifold with infinite injectivity radius, such as spheres, Kendall's shape spaces, and positive definite matrices with the afﬁne-invariant metric.

In practical application, it involves computation for vectors from different tangent spaces. However, directly performing these computations without considering the underline distinct tangent spaces may result in inaccurate outcomes. The core of our methodology is to transport the tangent vectors into a common tangent space, enabling direct computation such as vector subtraction and inner product in this common space. In the following, we will outline some properties associated with the parallel transformation operator on the tangent bundles of a manifold-valued curve.

\begin{definition}\label{proposition1}
Suppose that $A$, $B$, and $C$ are any three points on a measurable curve $g$ of $\mathcal{M}$, for any $v \in T_{A} \mathcal{M}$, it is moved by the parallel transportation $\Gamma_{A \rightarrow B}$ from $A$ to $B$ along $g$, denoted as $\Gamma_{A \rightarrow B}(v)$. Such parallel transport on a tangent bundle along the measurable curve $g$ on manifold $\mathcal{M}$ must satisfy,
\begin{itemize}
    \item $\Gamma_{A \rightarrow A}$ is an identity map;
    \item the dependence of  $\Gamma$  on $g$ is smooth;
    \item  $u=\Gamma_{A \rightarrow C}(v)=\Gamma_{B \rightarrow C}\Gamma_{A \rightarrow B}(v)$ and $
        v= \Gamma_{A \rightarrow C}^{-1}(u)=\Gamma_{A \rightarrow B}^{-1}\Gamma_{B \rightarrow C}^{-1}(u)$, where $\Gamma_{A \rightarrow B}^{-1}=\Gamma_{B \rightarrow A}$, $\Gamma_{A \rightarrow C}^{-1}=\Gamma_{C \rightarrow A}$ and $\Gamma_{B \rightarrow C}^{-1}=\Gamma_{C \rightarrow B}$.
\end{itemize}
\end{definition}

\begin{proposition}\label{proposition2}
Assuming that $A$ and $B$ are any two points on a measurable curve $g$ of $\mathcal{M}$, and $V_{A,1}, V_{A,2} \in T_{A} \mathcal{M}$ are any two different elements in the tangent space of $A$. For any $a,b \in \mathbb{R}$, we have
$\Gamma_{A \rightarrow B}(aV_{A,1}+bV_{A,2}) = a\Gamma_{A \rightarrow B}(V_{A,1})+b\Gamma_{A \rightarrow B}(V_{A,2})
$.
\end{proposition}

Definition \ref{proposition1} indicates that parallel transport along a smooth curve relies solely on the initial and final points, regardless of the specific path taken.  Proposition \ref{proposition2} demonstrates that the parallel transformation acts as a linear transformation, guaranteeing that random variables following a normal distribution still have the properties of normal distribution after transport. Examples of Riemannian manifolds: sphere manifold, SPD manifold and Grassmann manifold, are shown in Section 1 of the Supplementary Materials.

\subsection{Intrinsic covariance function of tangent space}

The wrapped Gaussian process regression model proposed in \cite{mallasto2018wrapped, liu2024wrapped} is to investigate the nonlinear probabilistic relationship between the response variables located on the manifold $\mathcal{M}$ and their corresponding covariates in a Euclidean space $\mathbb{R}^Q$. They assumed that the underlined manifold is isometrically embedded in a Euclidean ambient space, allowing for the processing of geometric objects such as tangent vectors within the ambient space. Although it is possible to account for some of the geometric structure in the ambient perspective, for example the curved nature of manifold via the Riemannian logarithm map, their methods do not address geometric properties across different tangent spaces and may have misleading results. We propose a new method to solve the problem by using a parallel transport.

\begin{figure}[ht!]
\vspace{-0.3cm}
    \centering
    \includegraphics[scale=0.5]{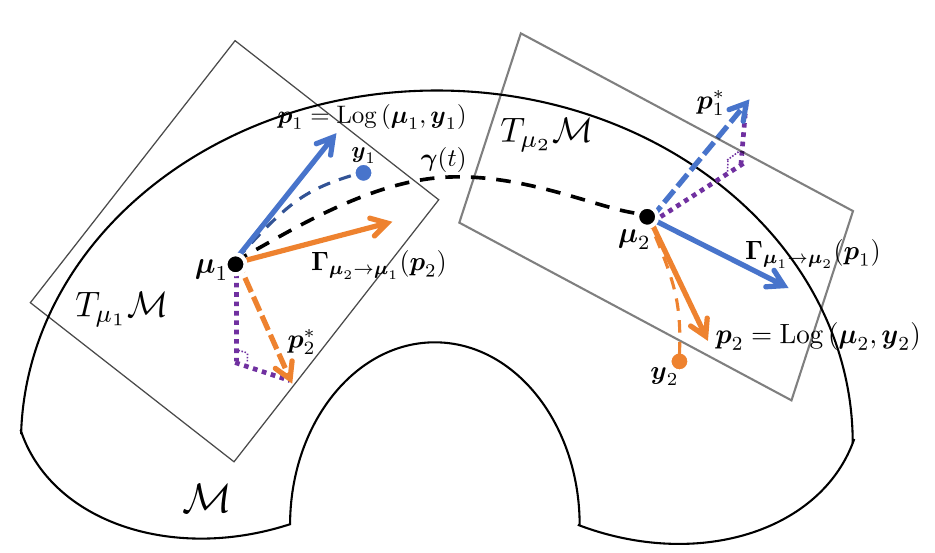}
     \caption{Ambient movement and parallel transport movement of the tangent vectors.}
     \label{FF1}
 \vspace{-0.5cm}
\end{figure}

To understand a parallel transport, consider the movements of two tangent vectors at different points as shown in Figure \ref{FF1}, where $p_1$ and $p_2$ are tangent vectors at the points $\mu_1$ and $\mu_2$ in the manifold $\mathcal{M}$, respectively. The curve $\gamma(t)$ is a geodesic from $\mu_1$ to $\mu_2$, $\gamma(0)=\mu_1$, $\gamma(1)=\mu_2$, and ${d\gamma(t)}/{dt}\mid_{t=0} = \operatorname{Log}(\mu_1,\mu_2)$.
The vector $p_1$, embeded in the ambient space, is moved along $\gamma(t)$ to the point $\mu_2$, while $p_2$ is similarly moved along $\gamma(t)$ to the point $\mu_1$. After the movements, we refer to the vectors as $p_1^{*}$ and $p_2^{*}$ respectively. Typically, $p_1^{*}$ and $p_2^{*}$ do not belong to the same tangent space. It follows that $p_1^{*}+p_2$ or $p_2^{*}+p_1$ is not tangent to the manifold either at $\mu_1$ or $\mu_2$. This discrepancy means that the movement in the ambient space can disrupt the intrinsic geometric information of the manifold-valued data. In other words, the direct computation (such as addition and inner product) of two tangent vectors without considering the underline distinct tangent spaces at different points is not an intrinsic geometric operation on the manifold. This may lead to errors from intrinsic geometry and potentially inappropriate covariance function if such a direct computation is used. Furthermore, since manifolds might be embedded into multiple ambient spaces, the interpretation of statistical results depends on the selected ambient space and could be misleading if the selection of ambient space is not made appropriately.

To solve the problem, we apply the parallel transport operator, a unique transport associated with the Riemannian metric, implemented through the Levi-Civita connection, to construct an intrinsic covariance function. The parallel transport operator $\Gamma_{\mu_1 \rightarrow \mu_2}: {T}_{\mu_1} \mathcal{M} \rightarrow$ ${T}_{\mu_2} \mathcal{M}$ intrinsically moves vectors between two different tangent spaces such that the inner products between two vectors before and after transport are conserved. Figure \ref{FF1} shows how the parallel transport operator works, where the transported tangent vectors, $\Gamma_{\mu_1 \rightarrow \mu_2}(p_1)$ and $p_2$, remain within the same tangent space along the curve. The parallel transport operator $\Gamma_{\mu_1 \rightarrow \mu_2}$  considers the geometric difference between the tangent spaces $T_{\mu_1} \mathcal{M}$ and $T_{\mu_2} \mathcal{M}$ when transporting the tangent vector $p_1$ to the tangent space $T_{\mu_2} \mathcal{M}$.


We now consider a Riemannian manifold $\mathcal{M}$ with an infinite injectivity radius. Let $\mu$ be a continuous measurable curve in the compact set $\mathbb{T}$, taking values in $\mathcal{M}$. Suppose $f$ is a manifold-valued random process on $\mathbb{T}$. For any $t,s,t_1,t_2 \in \mathbb{T}$, an intrinsic covariance structure can be defined by identifying a unique operator $\boldsymbol{\Phi}$ such that
\begin{equation}\label{r1}
\begin{aligned}
\boldsymbol{\Phi}_{\mu(t_1) \rightarrow \mu(t_2)}(\text{cov}(\Gamma_{\mu(t) \rightarrow \mu(t_1)}(\boldsymbol{\gamma}_t),\Gamma_{\mu(s) \rightarrow \mu(t_1)}(\boldsymbol{\gamma}_s))) \\
=\text{cov}(\Gamma_{\mu(t) \rightarrow \mu(t_2)}(\boldsymbol{\gamma}_t),\Gamma_{\mu(s) \rightarrow \mu(t_2)}(\boldsymbol{\gamma}_s)),
\end{aligned}
\end{equation}
where $\boldsymbol{\gamma}_t = \operatorname{Log}\left(\mu(t),y_t\right)$ and $\boldsymbol{\gamma}_s = \operatorname{Log}\left(\mu(s),y_s\right)$. The operator $\boldsymbol{\Phi}_{\mu(t_1) \rightarrow \mu(t_2)}$ be called the parallel transport operator of the covariance from the tangent space of $\mu(t_1)$ to the tangent space of $\mu(t_2)$ along the curve $\mu$.
Hence, we apply the parallel transport operator to propose a new intrinsic tangent space covariance function (iTSCF), which has the following property of invariance.

\begin{theorem}\label{t1}
Suppose that the operator $\boldsymbol{\Phi}_{\mu(t_1) \rightarrow \mu(t_2)}$ is defined in (\ref{r1}), there is a  
unique $\boldsymbol{\Phi}$ and the iTSCF: $\boldsymbol{\Phi}_{\mu(s) \rightarrow \mu(t)}(\text{cov}(\Gamma_{\mu(t) \rightarrow \mu(s)}(\boldsymbol{\gamma}_t),\boldsymbol{\gamma}_s))=\text{cov}(\boldsymbol{\gamma}_t,\Gamma_{\mu(s) \rightarrow \mu(t)}(\boldsymbol{\gamma}_s))$.
\end{theorem}

The proof of the existence and uniqueness of the operator $\boldsymbol{\Phi}$ is provided in Section 2 of the Supplementary Materials. Since random elements in the tangent space of the manifold are typically represented in the expressions of vectors or matrices \citep{calinon2020gaussians,zeestraten2017approach,lin2019intrinsic,zanini2016parameters}, a specific representation of the operator $\boldsymbol{\Phi}$ can be derived. In the following, we provide several examples of commonly used manifolds.

\begin{example}\label{example1}
{\bf Sphere manifold $\mathcal{S}^D$}. Suppose that $\mu(t)$ is a geodesic curve. The parallel transformation $\Gamma_{\mu(t) \rightarrow\mu(s)}(\boldsymbol{\gamma}_{t})$ is $\mathcal{A}_{\mu(t) \rightarrow\mu(s)}\boldsymbol{\gamma}_t$, where $\mathcal{A}_{\mu(t) \rightarrow\mu(s)}$ is a transformation matrix and $\boldsymbol{\gamma}_t \in T_{\mu(t)}\mathcal{S}^D$. Then, by Definition \ref{proposition1} and Proposition \ref{proposition2}, we have 
$\boldsymbol{\Phi}_{\mu(t_1) \rightarrow \mu(t_2)}(\cdot) = \mathcal{A}_{\mu(t_1) \rightarrow \mu(t_2)} (\cdot)\mathcal{A}^{\top}_{\mu(t_1) \rightarrow \mu(t_2)}$ and $\boldsymbol{\Phi}_{\mu(s) \rightarrow \mu(t)}(\cdot) = \mathcal{A}_{\mu(s) \rightarrow \mu(t)} (\cdot)\mathcal{A}^{\top}_{\mu(s) \rightarrow \mu(t)}$.
\end{example}

\begin{example}\label{example2}
{\bf SPD manifold $\mathcal{S}_{++}^D$ with affine-invariant metric}. Suppose that $\mu(t)$ is a geodesic curve. The parallel transformation $\Gamma_{\mu(t) \rightarrow\mu(s)}(\boldsymbol{\gamma}_{t})$ is $\mathcal{R}_{\mu(t) \rightarrow \mu(s)}\boldsymbol{\gamma}_{t}\mathcal{R}_{\mu(t) \rightarrow \mu(s)}^{\top}$, where $\boldsymbol{\gamma}_{t} \in T_{\mu(t)}\mathcal{S}_{++}^D$ is an asymmetric matrix, and $\mathcal{R}_{\mu(t) \rightarrow\mu(s)}=\mu(s)^{1/2}\mu(t)^{-1/2}$. Then, by Definition \ref{proposition1} and Proposition \ref{proposition2}, we have
$\boldsymbol{\Phi}_{\mu(t_1) \rightarrow \mu(t_2)}(\cdot) =(\mathcal{R}_{\mu(t_1) \rightarrow \mu(t_2)} \otimes\mathcal{R}_{\mu(t_1) \rightarrow \mu(t_2)}) (\cdot) (\mathcal{R}^{\top}_{\mu(t_1) \rightarrow \mu(t_2)} \otimes \mathcal{R}^{\top}_{\mu(t_1) \rightarrow \mu(t_2)})$ and $\boldsymbol{\Phi}_{\mu(s) \rightarrow \mu(t)}(\cdot) =(\mathcal{R}_{\mu(s) \rightarrow \mu(t)} \otimes\mathcal{R}_{\mu(s) \rightarrow \mu(t)} )(\cdot) (\mathcal{R}^{\top}_{\mu(s) \rightarrow \mu(t)} \otimes \mathcal{R}^{\top}_{\mu(s) \rightarrow \mu(t)})$, where $\otimes$ denotes the tensor product. 
\end{example}

The proofs of examples \ref{example1}-\ref{example2} are given in Section 3 of the Supplementary Materials.  For other manifolds, such as the hyperbolic manifold and the Grassmannian manifold \citep{calinon2020gaussians}, the parallel transformation $\boldsymbol{\Phi}$ can be derived similarly. 

By applying the iTSCF, we can build an intrinsic Gaussian process. 
Let $\operatorname{Log}({\vesub{\mu}{n}},\vesub{y}{n})=(\operatorname{Log}({\mu(t_1)}, y_{t_1}),
\cdots,\operatorname{Log}({\mu(t_n)}, y_{t_n})
)^\top$,
where $\vesub{\mu}{n}=\left(\mu\left(t_1\right), \ldots, \mu\left(t_n\right)\right)^\top$, and $\vesub{y}{n} = \left(y_{t_1}, \ldots, y_{t_n}\right)^\top$.
A random function $f$ with index in a set $\mathbb{T}$ and values in manifold $\mathcal{M}$ have an intrinsic  Gaussian process, if for any indices $t_1,\cdots,t_n$ over the set $\mathbb{T}$, vector $\vesub{f}{n} = \left(f\left(t_1\right), \ldots, f\left(t_n\right)\right)^\top$ follows 
\begin{equation}
\Gamma_{\vesub{\mu}{n} \rightarrow\mu(t)} \operatorname{Log}({\vesub{\mu}{n}},\vesub{f}{n}) = \boldsymbol{v}_{n}, ~~\boldsymbol{v}_{n} \sim \mathcal{N}(\mathbf{0}, K),\label{iwd}
\end{equation}
with 
covariance matrix $K = \{k_t(t_i, t_j)\}_{n\times n}$, denoted by 
\begin{equation}\label{e111}
f(t) \sim \mathcal{G}\mathcal{P}_{\mathcal{M}}(\mu(t), k_{t}(\cdot,\cdot)),\quad t \in \mathbb{T},
\end{equation} 
where $\mu(\cdot)$ is the basis point function (BPF) on a Riemannian manifold, and $k_{t}(\cdot,\cdot)$ defined in (\ref{r1})) is an intrinsic covariance function on a tangent space at $\mu(t)$.
In this paper, we call (\ref{iwd}) 
an intrinsic Gaussian distribution with the basepoint vector $\boldsymbol{\mu}_n$ and the covariance matrix $K$.

\vspace{-0.5cm}
\subsection{Intrinsic Gaussian process regression}\label{section2.3}
Using the intrinsic 
Gaussian process, we develop an intrinsic Gaussian process regression model (iGPR) for the manifold-valued response $y \in \mathcal{M}$ and the vector-valued predictor $\boldsymbol{x} \in \Omega$ via an unknown regression function $f: \Omega \rightarrow \mathcal{M}$. Let $\Omega = \mathbb{R}^Q$ first. Later, we will extend $\Omega$ to other spaces.
Given the training data $\mathcal{D}_{\mathcal{M}}:=\left\{\left(\boldsymbol{x}_{t_i}, y_{t_i}\right) \mid \boldsymbol{x}_{t_i} \in\right. \left.\mathbb{R}^Q, y_{t_i} \in \mathcal{M}, i=1, \cdots, n, t_i \in \mathbb{T} \right\}$, where $\mathbb{T}$ is a compact set, the main idea of the model is to assume an intrinsic Gaussian process as the prior of $f(\cdot)$. We consider ${T}_{\mu(t^{*})}\mathcal{M}$ as the common tangent space. We will show later that the estimation results are invariant to the choice of this common tangent space. The joint distribution between the training outputs $\boldsymbol{y}=\left(y_{t_1}, \cdots, y_{t_n}\right)$ and the test outputs $y_{t^*}$ at $\boldsymbol{x}_{t^*}$ is given by
\begin{equation}
\binom{y_{t^*}}{\boldsymbol{y}} \sim \mathcal{N}_{\mathcal{M}}\left(\binom{\mu^*}{\boldsymbol{\mu}},\left(\begin{array}{cc}
k_{t^*}\left(\boldsymbol{x}^*, \boldsymbol{x}^*\right) & K_{t^*}\left(\boldsymbol{X}, \boldsymbol{x}^*\right) \\
K_{t^*}^{\top}\left(\boldsymbol{X}, \boldsymbol{x}^*\right) & K_{t^*}(\boldsymbol{X}, \boldsymbol{X})
\end{array}\right)\right)
\end{equation}
where $\mu^*=\mu\left(t^{*}\right)$, $\boldsymbol{\mu}=\left(\mu\left(t_1\right), \ldots, \mu\left(t_{{n}}\right)\right)$, $ K_{t^*}(\boldsymbol{X}, \boldsymbol{X}) = \{k_{t^*}(\boldsymbol{x}_{t_i}, \boldsymbol{x}_{t_j})\}_{1 \leq i,j \leq n}$ and  $\boldsymbol{X}=\left(\boldsymbol{x}_{t_1}, \cdots, \boldsymbol{x}_{t_n}\right)$, $K_{t^*}\left(\boldsymbol{X}, \boldsymbol{x}^*\right) = (k_{t^*}\left(\boldsymbol{x}_{t_1}, \boldsymbol{x}^*\right),k_{t^*}\left(\boldsymbol{x}_{t_2}, \boldsymbol{x}^*\right),\cdots,k_{t^*}\left(\boldsymbol{x}_{t_n}, \boldsymbol{x}^*\right))$ and $k_{t^*}(\cdot,\cdot)$ is defined in (\ref{e111}), but with the common tangent space at $\mu(t^*)$.

In practice, we might need to work with a specific orthonormal bases for the tangent space. Therefore, we pick an orthogonal coordinate frame $\mathbf{E}$ for the common tangent space ${T}_{\mu(t^*)}\mathcal{M}$, and for any $t \in \mathbb{T}$, $\mathbf{E}_t = \Gamma_{\mu(t^*) \rightarrow \mu(t)}(\mathbf{E})$ is an orthogonal coordinate frame for the tangent space ${T}_{\mu(t)}\mathcal{M}$. Next, we will also show that the model estimation results are invariant to the choice of the orthogonal coordinate frame. Let $k_{\mathbf{E},t^*}$ be the representation of the intrinsic tangent space covariance function $k_{t^*}$ in the frame $\mathbf{E}$, and $\text{cov}_{\mathbf{E}}$ be the representation of the covariance function in frame $\mathbf{E}$. Thence, we have $k_{t^*}=\mathbf{E}^{\top}k_{\mathbf{E},t^*}\mathbf{E}$. And for any $t_1, t_2 \in \mathbf{T}$, we have $\text{cov}_{\mathbf{E}}(\Gamma_{\mu(t_1) \rightarrow \mu(t^*)}(\boldsymbol{\gamma}_{t_1}),\Gamma_{\mu(t_2) \rightarrow \mu(t^*)}(\boldsymbol{\gamma}_{t_2})) := k_{\mathbf{E},t^*}(\boldsymbol{x}_{t_1}, \boldsymbol{x}_{t_2})$.
Note that $k_{\mathbf{E},t^*}(\boldsymbol{x}_{t_1}, \boldsymbol{x}_{t_2})$ is the matrix-valued function. To obtain a valid $k_{\mathbf{E},t^*}(\boldsymbol{x}_{t_1}, \boldsymbol{x}_{t_2})$, a commonly used method is the intrinsic coregionalization model. This leads to
$
k_{\mathbf{E},t^*}(\boldsymbol{x}_{t_1}, \boldsymbol{x}_{t_2}) := k_{\mathbb{R}^{Q}}(\boldsymbol{x}_{t_1}, \boldsymbol{x}_{t_2},\boldsymbol{\theta}) \otimes \mathbf{B} \quad \text{with} \quad \mathbf{B} \in \mathbf{R}^{D \times D},
$
where $k_{\mathbb{R}^{Q}}$ is a scalar kernel function,  $\boldsymbol{\theta}$ is its kernel parameters, and $\mathbf{B}$ is a $D \times D$ coregionalization matrix that is positive semi-definite. 
Let matrix $K_{t^{*}} = (k_{\mathbb{R}^{Q}}(\boldsymbol{x}_{t_i}, \boldsymbol{x}_{t_j},\boldsymbol{\theta}))_{n\times n}$, and $\boldsymbol{k}_{t^*}=(k_{\mathbb{R}^{Q}}(\boldsymbol{x}_{t^*}, \boldsymbol{x}_{t_1},\boldsymbol{\theta}),k_{\mathbb{R}^{Q}}(\boldsymbol{x}_{t^*}, \boldsymbol{x}_{t_2},\boldsymbol{\theta}), \cdots , k_{\mathbb{R}^{Q}}(\boldsymbol{x}_{t^*}, \boldsymbol{x}_{t_n},\boldsymbol{\theta}))^{\top}$. 
Then predictive distribution of the test data $\boldsymbol{x}_{t^*}$ in tangent space of $\mu(t^{*})$ conditional on the training data $\mathcal{D}_{\mathcal{M}}$ is obtained as
\begin{equation}\label{ee1}
\operatorname{Log}(\mu(t^{*}),{y}(x_{t^{*}})) \mid \{\mathcal{D}_{\mathcal{M}},\boldsymbol{x}_{t^*},\boldsymbol{\mu},\mu^{*}\} \sim \mathcal{N}_{t^{*}}\left( {\boldsymbol{\beta}}_{t^{*}}^*, \boldsymbol{\Sigma}_{t^{*}}^*\right),
\end{equation}
where ${\boldsymbol{\beta}}_{t^*}^*= {\boldsymbol{\beta}}_{\mathbf{E}, t}^{*{\top}}{\mathbf{E}}$, ${\boldsymbol{\beta}}_{\mathbf{E}, t}^*=\boldsymbol{K}_{\mathbf{E},t^{*}}^* \boldsymbol{K}_{\mathbf{E},t^{*}}^{-1}(\Gamma_{\boldsymbol{\mu} \rightarrow \mu_{t^{*}}}(\operatorname{Log}(\boldsymbol{\mu}, \boldsymbol{y})))_{\mathbf{E}}$, $\boldsymbol{\Sigma}_{t^{*}}^*={\mathbf{E}}^{\top}\overline{\boldsymbol{\Sigma}}^{*}_{\mathbf{E},t}{\mathbf{E}}$,  $\overline{\boldsymbol{\Sigma}}^{*}_{\mathbf{E},t}=\boldsymbol{K}_{\mathbf{E},t^{*}}^{* *}-\boldsymbol{K}_{\mathbf{E},t^{*}}^* \boldsymbol{K}_{\mathbf{E},t^{*}}^{-1} \boldsymbol{K}_{\mathbf{E},t^{*}}^{* \top}$, $\boldsymbol{K}_{\mathbf{E},t^{*}}^*=\mathbf{B}\otimes\boldsymbol{k}_{t^*}$, $\boldsymbol{K}_{\mathbf{E},t^{*}}=\mathbf{B}\otimes K_{t^{*}}$ and $\boldsymbol{K}_{\mathbf{E},t^{*}}^{* *} = k_{\mathbb{R}^{Q}}(\boldsymbol{x}_{t^*}, \boldsymbol{x}_{t^*};{\boldsymbol{\theta}}) \otimes \mathbf{B}$. 
One can see that $(\Gamma_{{\mu}(t_i) \rightarrow \mu(t^{*})}(\operatorname{Log}({\mu}(t_i),{y}_{t_i})))_{\mathbf{E}}$ is a coordinate representation of $\Gamma_{{\mu}(t_i) \rightarrow \mu(t^{*})}(\operatorname{Log}({\mu}(t_i),{y}_{t_i}))$, and 
$
(\Gamma_{\boldsymbol{\mu} \rightarrow \mu_{t^{*}}}(\operatorname{Log}(\boldsymbol{\mu}, \boldsymbol{y})))_{\mathbf{E}}=\text{Vec}(((\Gamma_{{\mu}(t_1) \rightarrow \mu(t^{*})}(\operatorname{Log}({\mu}(t_1),{y}_{t_1})))_{\mathbf{E}}, \cdots , (\Gamma_{{\mu}(t_n) \rightarrow \mu(t^{*})}(\operatorname{Log}({\mu}(t_n), {y}_{t_n})))_{\mathbf{E}})^{\top}).
$

It follows that $\operatorname{Exp}(\mu(t^*),{\boldsymbol{\beta}}_{t^*}^*)$ can be used as a point prediction of $y_{t^*}$. Although the predictive distribution $\mathcal{N}_{t^{*}}$ is not necessarily intrinsic Gaussian distribution, as $\boldsymbol{\beta}_{t^{*}}^*$ might be non-zero. As in \cite{hong2017regression}, the distribution can be approximated using a intrinsic Gaussian distribution  with the basepoint $\operatorname{Exp}_{{\mu}(t^*)}\left(\boldsymbol{\beta}_{t^{*}}^*\right)$ and parallel transporting the tangent space covariance $\boldsymbol{\Sigma}_{t^{*}}^*$ to this basepoint along the corresponding geodesic. In addition, these hyperparameters $\boldsymbol{\theta}$ and $\mathbf{B}$ can be obtained by maximizing the logarithmic marginal likelihood. As there is no analytic solution, the recursive algorithm developed by \cite{cao2023gaussian,boumal2014manopt} can be used. A detailed computation procedure for iGPR is proved in Algorithm \ref{Algorithm1}.

\begin{algorithm}
    \caption{iGPR}
    \renewcommand{\algorithmicrequire}{\textbf{Input:}}
    \renewcommand{\algorithmicensure}{\textbf{Output:}}
    \label{Algorithm1}
    \begin{algorithmic}[1]
        \REQUIRE $\left\{\left(\boldsymbol{x}_{t_i}, y_{t_i}\right)\right\}_{i=1}^n$, $\{t_i\}_{i=1}^{n}$, $t^*$, $\boldsymbol{x}^*$, the basepoint function $\mu(t)$ and the kernel function $k_{\mathbb{R}^{Q}}$;
        
        \STATE Select an orthonormal frame $\mathbf{E}$ at the tangent space of $\mu(t^*)$.
        
        \FOR{all $t_i, i=1,2,\cdots,n$}
            \STATE Compute the orthonormal frames $\{\mathbf{E}_{t_i}\}_{i=1}^n$ using $\mathbf{E}_{t_i} = \Gamma_{\mu(t^*) \rightarrow \mu(t_i)}(\mathbf{E})$;
        \ENDFOR
        
        \STATE Calculate the $\mathbf{E}$-coordinate representations of $\Gamma_{\mu(t_i) \rightarrow \mu_{t^*}}(\operatorname{Log}(\mu(t_i), y_i))$, $i=1,2,\cdots,n$;
        
        \STATE Estimate hyperparameters $\boldsymbol{\theta}$ and $\mathbf{B}$ via maximum log marginal likelihood estimation;
        
        \STATE Compute the predictive distribution $\mathcal{N}_{t^{*}}\left( {\boldsymbol{\beta}}_{t^{*}}^*, \boldsymbol{\Sigma}_{t^{*}}^*\right)$ based on (\ref{ee1});
        
        \ENSURE the maximum a posteriori (MAP) estimate $\operatorname{Exp}(\mu(t^*), {\boldsymbol{\beta}}_{t^*}^*)$.
    \end{algorithmic}
\end{algorithm}

Next, we need to address the problem of how to choose the common tangent space. The process so far has been carried out in the tangent space of $\mu(t^*)$. What would happen if we were to perform the analysis in the tangent space of other points along the BPF ? Theorem \ref{theorem1} below provides an answer: the posterior distribution $\mathcal{N}_{t}$ under the tangent space of other points $\mu(t)$ on the BPF, after applying a parallel transformation $\boldsymbol{\Phi}_{\mu(t) \rightarrow \mu(t^*)}$ along the BPF, remains consistent with the posterior distribution $\mathcal{N}_{t^*}$ in the tangent space of $\mu(t^*)$. The proof is given in Section 4 of the Supplementary Materials.

\begin{theorem}\label{theorem1}
    Assuming that the BPF $\mu(t)$ is a smooth curve. For any $t \in \mathbf{T}$, let $\mathbf{A}$ be a frame for the tangent space of $\mu(t)$ and $\Gamma_{\mu(t) \rightarrow \mu(t^*)}(\mathbf{A})=\mathbf{E}$. Then, we have
$
    {\boldsymbol{\beta}}_{t^*}^*=\Gamma_{\mu(t) \rightarrow \mu(t^{*})}({\boldsymbol{\beta}}_{t}^*); \quad
    \boldsymbol{\Sigma}_{t^*}^* = \boldsymbol{\Phi}_{\mu(t) \rightarrow \mu(t^{*})}(\boldsymbol{\Sigma}_{t}^*),  
$
where ${\boldsymbol{\beta}}_{t}^*$ and $\boldsymbol{\Sigma}_{t}^*$ represent the mean and variance of the predicted distribution $\mathcal{N}_{t}$ in the tangent space of $\mu(t)$.
\end{theorem}

Furthermore, the following Theorem \ref{theorem2} demonstrates that the predicted distribution is independent of the chosen frame. The proof is given in Section 5 of the Supplementary Materials.

\begin{theorem}\label{theorem2}
For any $t \in \mathbf{T}$, suppose $\mathbf{W}$ is another frame on the tangent spaces of $\mu(t)$. Change from $\mathbf{E}$ to $\mathbf{W}$ can be characterized by a unitary matrix $\mathbf{O}$, and $\mathbf{W}=\mathbf{O}^{\top}\mathbf{E}$. Then, we have
$
{\boldsymbol{\beta}}_{\mathbf{E}, t}^* = \mathbf{O}{\boldsymbol{\beta}}_{\mathbf{W}, t}^*; \quad \overline{\boldsymbol{\Sigma}}^{*}_{\mathbf{E},t}   = \mathbf{O}\overline{\boldsymbol{\Sigma}}^{*}_{\mathbf{W},t}\mathbf{O}^{\top}.
$
This is, the mean and variance of the predicted distribution is independent of the frame.
\end{theorem}

\begin{figure}[!ht]
    \centering
    \includegraphics[scale=0.5]{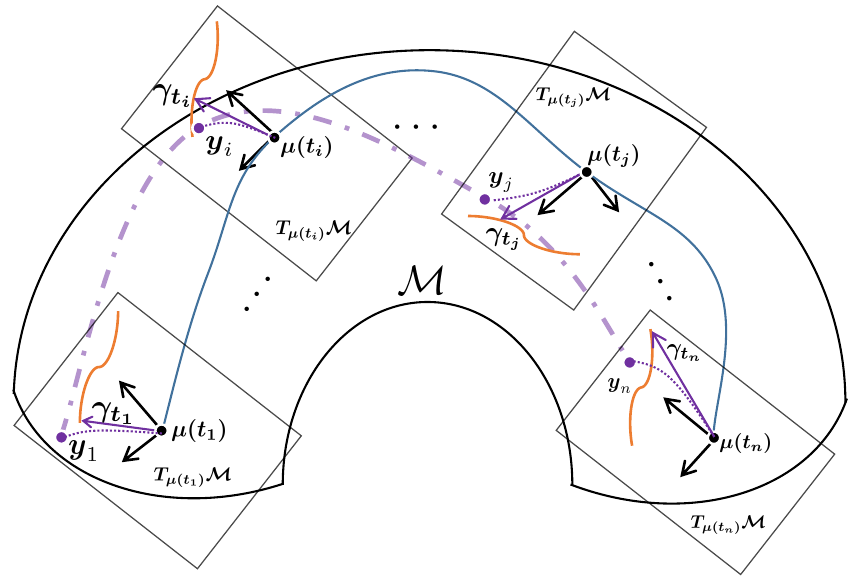}
    \caption{
    Logarithm map of the manifold-valued data and schematic diagram of equivalent version of iGPR, where the blue solid line represents BPF.}
     \label{FF2}
\end{figure}

\vspace{-0.5cm}
The proposed iGPR uses the curvature information of the manifold by implementing parallel transport from different tangent spaces along the BPF $\mu(t)$ to a common tangent space, affectively merging the intrinsic geometric structure of the manifold. We have shown that the estimation for the iGPR model is unaffected by the selection of coordinate frames or common tangent space. In addition, we find an equivalent version of the proposed iGPR: parallel transporting the coordinate frame along the BPF results in a series of coordinate frames across different tangent spaces, leading to a representation of random elements in these frames within a Euclidean space $\mathbb{R}^{D}$. We also show that these representations retain consistency despite changes in the tangent spaces. Figure \ref{FF2} gives a geometric illustration, where the black arrows represent coordinate frames obtained via parallel transport operator on different tangent spaces, and the orange solid lines stand for the representation of random elements after a parallel transport from the common tangent space to different tangent spaces. This results in the purple dashed lines connecting $y_i$, which is the curve predicted from the intrinsic Gaussian process regression model. The relative position between the orange solid line and coordinate axes in black remains unchanged, indicating that the choice of different common tangent space does not affect the model estimation. After representing the random elements of different tangent spaces using the aforementioned coordinate frame, they reside within the same Euclidean space and can be modeled by a vector-valued Gaussian process regression model.

We assumed a Euclidean predictor so far. The proposed method, however, can be extended to a broader framework that can model response variables residing on manifolds with mixed Euclidean or non-Euclidean predictors. 
\begin{remark}\label{remark1}
By incorporating kernel functions specifically designed for non-Euclidean predictors within the iGPR model, our proposed method still works. Let the predictor $\boldsymbol{x} \in \Omega$, where $\Omega$ is a general space such as a metric space. As long as there is a kernel function $k_{\Omega}$ with symmetric and positive definite properties, the estimation and prediction computations are carried out similarly to those in Section 2.3 while
$k_{\mathbb{R}^{Q}}$ is replaced with $k_{\Omega}$. For example, when $\Omega$ is a Riemannian manifold, one might select either the heat kernel \citep{niu2019intrinsic,niu2023intrinsic} or the Matérn kernel \citep{borovitskiy2021matern}. When $\Omega$ is a general metric space, the options include the graph kernel \citep{dunson2022graph,vishwanathan2010graph} and the persistence weighted Gaussian kernel \citep{kusano2016persistence}.
\end{remark}

\subsection{iGPR with observation noise}
We didn't consider noise in the model discussed in Section \ref{section2.3}. In practical applications, data typically include random fluctuations and observational errors. By incorporating noise into the iGPR framework, the model can more accurately distinguish between signal and noise and then capture the underlying patterns. Furthermore, the inclusion of noise guarantees the positive definiteness of the covariance matrix, ensuring numerical stability. The iGPR model with error terms can be defined as
\begin{equation}\label{q14}
y(x_t):=\operatorname{Exp}\left(\mu(t), \boldsymbol{\gamma}_t(x_t) +\epsilon_t \right)
\end{equation}
where $\epsilon_t$'s are independent measurement errors following a multivariate normal distribution in the tangent space of $\mu(t)$, i.e, $\epsilon_t \sim \mathcal{N}\left(0, \sigma^2\mathbf{E}_t^{\top}\boldsymbol{I}_{D \times D}\mathbf{E}_t\right)$ , here $\mathbf{E}_t$ is a frame for the tangent space ${T}_{\mu(t)}\mathcal{M}$ and $\mathbf{E}_t = \Gamma_{t^{*} \rightarrow t}(\mathbf{E})$.

Inclusion of this error changes the regression procedure only slightly. Following similer derives in the previous subsections, we have the joint distribution of $\boldsymbol{y}$ and $y_{t^*}$ as
{\scriptsize{
$$
\binom{y_{t^*}}{\boldsymbol{y}} \sim \mathcal{N}_{\mathcal{M}}\left(\binom{\mu_{t^*}}{\boldsymbol{\mu}},\left(\begin{array}{cc}
\mathbf{E}^{\top}\boldsymbol{K}_{\mathbf{E},t^{*}}^{* *}\mathbf{E}, & \mathbf{E}^{\top}\boldsymbol{K}_{\mathbf{E},t^{*}}^*(\mathbf{E} \otimes \boldsymbol{I}_{n \times n}) \\
(\mathbf{E}^{\top}\boldsymbol{K}_{\mathbf{E},t^{*}}^*(\mathbf{E} \otimes \boldsymbol{I}_{n \times n}))^\top, & (\mathbf{E} \otimes \boldsymbol{I}_{n \times n})^{\top} (\boldsymbol{K}_{\mathbf{E},t^{*}}+\sigma^2\boldsymbol{I}_{nD \times nD}) (\mathbf{E} \otimes \boldsymbol{I}_{n \times n})\end{array}\right)\right),
$$}
}where $\mathcal{N}_{\mathcal{M}}$ is the intrinsic Gaussian distribution. The remaining procedure for computing estimation and prediction is the same as the one in Subsection \ref{section2.3}, where $(\mathbf{E} \otimes \boldsymbol{I}_{n \times n})^{\top} \boldsymbol{K}_{\mathbf{E},t^{*}} (\mathbf{E} \otimes \boldsymbol{I}_{n \times n}) $ is replaced with $(\mathbf{E} \otimes \boldsymbol{I}_{n \times n})^{\top}(\boldsymbol{K}_{\mathbf{E},t^{*}} +\sigma^2\boldsymbol{I}_{nD \times nD})(\mathbf{E} \otimes \boldsymbol{I}_{n \times n}) $. Theorem \ref{theorem3} below proves information consistency of the prediction obtained by the model (\ref{q14}). Suppose that the underlying true vector field $\boldsymbol{\gamma}(t)$ along the BPF $\mu(t)$ can be represented by real-valued function vectors $\boldsymbol{\gamma}_{\mathbf{E}}$ based on an orthogonal coordinate frame $\{\mathbf{E}_t\}_{t \in \mathbf{T}}$, where $\Gamma_{t \rightarrow s}(\mathbf{E}_t)=\mathbf{E}_s$ for any $t,s \in \mathbf{T}$. Without loss of generality, we consider a special setting of $k_{\mathbf{E},t^*}$, i.e. the coordinates are independent of each other, i.e. $k_{\mathbf{E},t^*}\left(\cdot, \cdot\right)=\text{diag}(k_{\mathbb{R}^{Q}}\left(\cdot, \cdot, \boldsymbol{\theta}_1\right),\cdots,k_{\mathbb{R}^{Q}}\left(\cdot, \cdot, \boldsymbol{\theta}_D\right))$. Then, assume that the random element $\{\boldsymbol{\gamma}_{\mathbf{E},d}\},d=1,2,\cdots,D$ has a Gaussian process prior with mean zero and a
bounded covariance function $ k_{\mathbb{R}^{Q}}\left(\cdot, \cdot, \boldsymbol{\theta}_d\right)$ for any covariate $\boldsymbol{x}_{t}$. Suppose the $k_{\mathbb{R}^{Q}}\left(\cdot, \cdot, \boldsymbol{\theta}_d\right)$ is continuous in $\boldsymbol{\theta}_d$, further assume the estimator $\hat{\boldsymbol{\theta}}_d$ converges to $\boldsymbol{\theta}_d$ as $n \rightarrow \infty$ almost surely, and the BPF $\mu(t)$ is known.

\begin{theorem}\label{theorem3}
Under the above assumptions, let the reproducing kernel Hilbert space norm be bounded and the expected regret term $E\left(\log \left|\boldsymbol{I}_{n \times n}+\sigma^{-2} \boldsymbol{K}_{d}\right|\right)=o(n)$ for $d=1,\cdots,D$, where $\boldsymbol{K}_{d}=\left(k_{\mathbb{R}^Q}\left(\boldsymbol{x}_{t_i}, \boldsymbol{x}_{t_j}; \boldsymbol{\theta}_d\right)\right)_{1 \leq i,j \leq n}$ is the $d$-dimensional representation of the covariance matrix over $\{\boldsymbol{x}_{t_i}\}_{i=1}^{n}$ under the orthogonal coordinate frames $\{\mathbf{E}_t\}_{t \in \mathbf{T}}$, and $\sigma^2$ is the variance of the measurement error. Then, the estimator $\hat{\boldsymbol{\gamma}}_{\mathbf{E},d}$ is information consistent to the true ${\boldsymbol{\gamma}}_{\mathbf{E},d}$, which means an estimator that guarantees convergence to the true parameter value as the sample size grows.
\end{theorem}

The proof is given in Section 6 of the Supplementary Materials. Furthermore, we also consider the consistency of the posterior distribution. This is shown in Theorem \ref{theorem4} and the proof is given in Section 7 of the Supplementary Materials.

\begin{theorem}\label{theorem4}
Suppose that the underlying true vector field $\boldsymbol{\gamma}(t)$ along the BPF $\mu(t)$ can be represented by real-valued function vectors $\boldsymbol{\gamma}_{\mathbf{E}}$ based on an orthogonal coordinate frame $\{\mathbf{E}_t\}_{t \in \mathbf{T}}$, where $\Gamma_{t \rightarrow s}(\mathbf{E}_t)=\mathbf{E}_s$ for any $t,s \in \mathbf{T}$. Without loss of generality, we consider that the coordinates are independent of each other and that the BPF $\mu(t)$ is known. Let ${\boldsymbol{\gamma}}_{\mathbf{E},d}: \mathbb{R}^{Q} \rightarrow \mathbb{R}$ be a function and $\widehat{v}_{n,d}(\boldsymbol{x}_t)$ denote the variance of the posterior distribution of the estimator $\hat{\boldsymbol{\gamma}}_{E,d}$. Then, for every $d =1,2,\cdots,D$, we have
$
\sup _{\boldsymbol{x} \in \mathbb{R}^{Q}} \widehat{v}_{n,d}(\boldsymbol{x}_t) \longrightarrow 0, \quad n \rightarrow \infty,
$
almost surely monotonically, as well as in $L^p$ for every $p \in[1, \infty)$. Furthermore, if ${\boldsymbol{\gamma}}_{\mathbf{E},d}$ lies in the RKHS of $k_{\mathbb{R}^{Q}}\left(\cdot, \cdot, \boldsymbol{\theta}_d\right)$, for all $\boldsymbol{x}_t \in \mathbb{R}^{Q}$ and $d =1,2,\cdots,D$, we have
$
\hat{\boldsymbol{\gamma}}_{\mathbf{E},d}(\boldsymbol{x}_t) \xrightarrow{L^2} {\boldsymbol{\gamma}}_{\mathbf{E},d}(\boldsymbol{x}_t), \quad n \rightarrow \infty.
$
\end{theorem}

\begin{remark}
In Theorem \ref{theorem4}, it needs the condition that $\mu(t)$ is known. But in practice, $\mu(t)$ is usually unknown and is estimated using data. In this case, we have the following conclusion. Consider the same conditions as in Theorem \ref{theorem4} and suppose $\sup _{t \in \mathcal{T}} d_{\mathcal{M}}^2(\hat{\mu}(t), \mu(t))=O_P\left(n^{-1}\right)$ holds. Then, we have $(\Gamma_{\hat{\mu}(t)\rightarrow\mu(t)}(\hat{\boldsymbol{\gamma}}(\boldsymbol{x}_t)))_{\mathbf{E},d} \xrightarrow{L^2} {\boldsymbol{\gamma}}_{\mathbf{E},d}(\boldsymbol{x}_t)$, for all $\boldsymbol{x}_t \in \mathbb{R}^{Q}$, $n \rightarrow \infty$. Furthermore, from Theorem 6 of \cite{lin2019intrinsic}, assuming that the basepoint function is an intrinsic mean curve, condition $\sup _{t \in \mathcal{T}} d_{\mathcal{M}}^2(\hat{\mu}(t), \mu(t))=O_P\left(n^{-1}\right)$ holds.
\end{remark}

\section{Simulation studies}\label{s2}
In this section, we conduct several numerical experiments to illustrate the performance of the proposed iGPR model. We consider sample sizes of $N=30,50,100,150$ and two setups. The first is a random sampling scheme (Random), where 80\% of the $N$ observation points are randomly selected as training data and the remaining are the test data. The second setup is a sorted sampling scheme (Sort), in which the last five samples of the $N$ sample points are designated
as the test data and the remainings are the training data. All simulation results are based on 100 independent repetitions, and the average and standard errors of the root-mean-square geodesic errors are computed. To evaluate the performance of the proposed method, we compare it with wrapped Gaussian process regression (WGPR) \citep{liu2024wrapped} and vector-valued Gaussian process regression (MGPR) \citep{alvarez2012kernels}.

This section considers an example: the symmetric positive definite matrices space $\mathcal{S}_{++}^3$ with affine-invariant metric, while another example: the SPD manifold on the unit sphere, is presented in Section 9 of the Supplementary Materials. Let $\mu(t)$ be a geodesic curve $\mu(t) = \operatorname{Exp}(\mu(0),\operatorname{Log}(\mu(0),(2.4360,$ $-2.6465;-2.6465,8.4208)^{\top})t),$ 
where $\mu(0)=(1,0;0,1)$, the time domain is $t \in [0,1]$ and the predictor is time, i.e, $\boldsymbol{x}_{t}=t$. The initial frame $(1 ,0;0,0)$, $(0,1;1,0)$ and $(0,0;0,1)$ are utilized in the tangent space of $\mu(0)$. Let the coordinates be independent using the tangle space of $\mu(0)$ as initial tangle space to generate the covariance function. 
We consider the following two scenarios. (\textbf{S1}) We use the following Gaussian kernel function matrix as the intrinsic covariance function under the frame $\mathbf{E}$,
$$ 
k_{\mathbf{E},t^*}(\boldsymbol{x}, \tilde{\boldsymbol{x}}) := \begin{pmatrix}
 \exp({\frac{-\|\boldsymbol{x} - \tilde{\boldsymbol{x}}\|_{2}^2}{\theta_{j1}}}) & 0 & 0 \\
 0 & \exp({\frac{-\|\boldsymbol{x} - \tilde{\boldsymbol{x}}\|_{2}^2}{\theta_{j2}}}) & 0 \\
 0 & 0 & \exp({\frac{-\|\boldsymbol{x} - \tilde{\boldsymbol{x}}\|_{2}^2}{\theta_{j3}}})
\end{pmatrix}, \boldsymbol{x}, \tilde{\boldsymbol{x}} \in [0,1]; j=1,2,3,
$$
with parameters $(\theta_{11},\theta_{12},\theta_{13})=(0.1,~0.3,~0.5)^\top$, $(\theta_{21},\theta_{22},\theta_{23})=(0.3,~0.6,~0.8)^\top$ and $(\theta_{31},\theta_{32},\theta_{33})=(0.5,~0.8,~0.8)^\top$. (\textbf{S2})  $(\gamma_{1,\mathbf{E}}(t), \gamma_{2,\mathbf{E}}(t),\gamma_{3,\mathbf{E}}(t))$ is used as the vector-valued function under the frame $\mathbf{E}$, where $\gamma_{1,\mathbf{E}}(t)=1 +0.05*t + \sin(t)/(t+0.001) + 0.01*\epsilon_{1,t}$ , $\gamma_{2,\mathbf{E}}(t)=1+\cos(t)+0.03*\epsilon_{2,t}$ and $\gamma_{3,\mathbf{E}}(t)=1+\sin(t)+0.05*\epsilon_{3,t}$. Here, $\{\epsilon_{l,t}\}_{l=1}^{3}$ are independent standard normal distributions.

\begin{table}[!ht]
\vspace{0.1cm}
\tabcolsep=5pt\fontsize{10}{16}\selectfont
\centering
\caption{The average and standard errors of the root mean square geodesic errors of WGPR, iGPR and MGPR on the SPD manifold $\mathcal{S}_{++}^3$ with \textbf{S1}.}
\label{table3}
\begin{tabular}{cccccc ccccc c}
\hline
 & \multirow{1}{*}{Scheme} &  \multirow{1}{*}{$\boldsymbol{\theta}$}  & $n$  & \multicolumn{1}{c}{iGPR} & \multicolumn{1}{c}{WGPR} & \multicolumn{1}{c}{MGPR} \\\hline
 &Random & (0.1,0.3,0.5) & 30               &  0.0400(0.1037)    &0.5841(3.7246)                 &4.1920(0.5894)  \\
 && & 50              & 0.0154(0.0939)   & 0.2106(1.5451)                  & 2.6372(0.3849) \\
&& & 100               & 0.0051(0.0082)   & 0.0686(0.0568)                   &3.5720(0.5291) \\
&& & 150               &0.0025(0.0034)  &0.0166(0.0132)                       &2.9938(0.4472) \\
& & (0.3,0.6,0.8)  &30               & 0.0118(0.0338)    &0.4822(1.4677)                &4.4950(0.5229)      \\
& & & 50                 & 0.0101(0.0775) & 0.1396(0.1231)                   &4.0192(0.6984)    \\
&& & 100                & 0.0026(0.0048)  & 0.0606(0.0718)                    &3.5103(0.4792)     \\
&& & 150               & 0.0018(0.0027)  & 0.0301(0.0683)                 &3.6349(0.4869)     \\ 
& & (0.5,0.8,0.8)  &30               & 0.0084(0.0154)     &0.5669(1.3977)                    &4.8021(0.7293)       \\
& & & 50                 & 0.0054(0.0135)  & 0.3837(1.5958)                    &3.9927(0.4038)    \\
&& & 100                & 0.0035(0.0062) & 0.0875(0.1712)                     & 3.4932(0.5282)       \\
&& & 150                & 0.0012(0.0015) & 0.0183(0.0244)                     & 3.1196(0.4389)    \\ 

&Sort & (0.1,0.3,0.5) & 30                &0.6911(0.9584)    &2.7429(3.2810)                    & 5.8930(0.3920)     \\
&& & 50                  &0.2234(0.6580)       &1.5734(1.9173)                  & 5.5487(0.5593)    \\
& & & 100                & 0.0655(0.1130)        &0.3289(0.7900)                & 4.4905(0.4939)  \\
& &  & 150                   & 0.0419(0.0712)       & 0.1619(0.3433)                 &3.5594(0.7392)  \\
& & (0.3,0.6,0.8) & 30                & 0.2739(0.5176)        &2.2021(3.0825)                  &5.9368(0.4473)   \\
& &  &50                    & 0.1336(0.1465)    & 1.0074(1.1349)                &4.1209(0.4850)    \\
& &  &100                     & 0.0265(0.0467)    & 0.2164(0.5824)                 &4.4639(0.5948)  \\
& &  & 150                     &0.0149(0.0245)    & 0.1547(0.6929)                   &4.6392(0.6483)    \\ 
& & (0.3,0.6,0.8) & 30                & 0.1535(0.2747)     &1.9174(3.1997)                  &5.0020(0.5684) \\
& &  &50                      & 0.0758(0.0968)     & 1.3795(1.0016)                 &4.4904(0.3899)  \\
& &  &100                    & 0.0224(0.0378)      & 0.4189(0.8489)                  &4.3992(0.5289)   \\
& &  & 150                     &0.0146(0.0278)       & 0.1927(0.6689)                & 3.1192(0.5859)   \\ \hline
\end{tabular}
\vspace{-0.5cm}
\end{table}

Average values and standard errors of the root mean square geodesic errors from three methods, iGPR, WGPR and MGPR, are shown in Tables \ref{table3}-\ref{table4}, corresponding to scenarios \textbf{S1} and \textbf{S2} respectively. From both Tables, we see that iGPR has the smallest root mean square geodesic errors, indicating that iGPR has a significant improvement in predictive performance compared to WGPR and MGPR. In addition, MGPR has the largest root mean square geodesic errors, as it fails to account for the geometric information inherent in manifold-valued data during modeling, thus violating the intrinsic structure of the manifold. Although WGPR incorporates the local nonlinear structure of the manifold through exponential and logarithmic maps, it still disrupts the manifold's intrinsic geometric structure by directly computing tangent vectors from different tangent spaces during modeling. As the sampling points become sparser, the proposed method still maintains excellent performance, whereas the performance of WGPR declines significantly. 

We also investigate the ability of the proposed method to reconstruct the covariance function of Gaussian process. The results are shown in Section 8 of the Supplementary Materials, which illustrates the estimated covariance function in the tangent space compared to the true covariance function with kernel parameters for covariance structure reflecting the mean values from 100 experiments. The heatmaps indicate that the estimated covariance functions align more closely with the true one compared with WGPR.  Additionally, similar experiments were conducted on the unit sphere, and the results are generally consistent with those obtained on the SPD manifold. For more details, see Section 9 of the Supplementary Materials.

\begin{table}
\vspace{0.1cm}
\tabcolsep=5pt\fontsize{10}{16}\selectfont
\centering
\caption{The average and standard errors of the root mean square geodesic errors of WGPR, iGPR and MGPR on the SPD manifold $\mathcal{S}_{++}^3$ with \textbf{S2}.}
\label{table4}
\begin{tabular}{cccccc ccccc c}
\hline
 & \multirow{1}{*}{Scheme} &  \multirow{1}{*}{$\boldsymbol{\sigma^2}$}  & $n$  & \multicolumn{1}{c}{iGPR} & \multicolumn{1}{c}{WGPR} & \multicolumn{1}{c}{MGPR} \\\hline
& Random & (0.001,0.003,0.005) & 30   & 0.3468(0.0211)  &0.6069(0.0429)        & 4.3950(0.7493)    \\
& & & 50          & 0.1918(0.0139) &0.5403(0.0424)        &4.0192(0.8377)     \\
& & & 100                             & 0.0957(0.0192)  &0.4236(0.0402)        &4.3302(0.8849)    \\
& & & 150                           & 0.0135(0.0491)  &0.0402(0.4320)        &4.7621(0.9010)     \\
&Sort & (0.001,0.003,0.005) & 30                & 0.9026(1.1006) & 1.3051(1.0743)         &5.3238(1.2947)       \\
& & & 50                  & 0.5210(0.7340)  & 0.9412(0.7440)       &4.8392(0.8848)   \\
& & & 100                 & 0.1435(0.4970)  & 0.3474(0.5010)       &4.9793(0.7904) \\
& &  & 150                &0.0614(0.4287) & 0.2655(0.4320)      &5.5436(0.7694)       \\ \hline
\end{tabular}
\vspace{-0.5cm}
\end{table}

\section{Real data analysis}\label{section:4}
\subsection{Flight trajectory data}
The earth is roughly a sphere and can be modelled as a copy of ${\mathcal{S}^{2}}$, and the flight trajectory can be therefore considered as a manifold-valued random curve. In this article, the flight trajectory data is from the FlightAware website (\url{https://www.flightaware.com/}). Specifically, the flight trajectory data used in this study come from the flight tracking and historical data of Cathay Pacific Airways. These flight trajectories of 20 flights from Hong Kong to London in the period of Jul 30, 2024-Aug 10, 2024. To obtain smooth curves from the occasionally noisy data, we pre-smoothed the longitude and latitude data using kernel local linear smoothing and then mapped the longitude and latitude trajectories onto a unit sphere $\mathcal{S}^2$.

We select the position of the airplane as the response variable, which can be represented on $S^2$ using longitude and latitude, with time considered as covariates. We set the takeoff time of the flight as 0 and the landing time as 1. To account for different levels of sparsity, we consider total data points of $N=30$, $50$, and $100$, chosen at equal intervals from the flight trajectory data. Similar to the previous section, we consider two setups: (1) randomly selecting 20\% of the $N$ observation points recorded on August 10, 2024 as test data, and the others as training data (Random); (2) the last five points of the $N$ observation points in each trajectory as test data, and the others as training data (Sort). For the Random setup, the procedure is repeated independently 100 times, and for the Sort setup the 20 flight trajectories are computed separately by using the proposed method. These procedures yield the evaluation metrics: mean and standard errors of the root mean square geodesic errors. Geodesic regression estimation is used as the prior BPF, and a diagonal RBF kernel is taken as the prior TSCF. The prediction accuracies of iGPR, WGPR and MGPR on the flight trajectory data are presented in Table \ref{table5}. We see that the iGPR model has the smallest estimation errors, and outperforms both WGPR and MGPR in terms of interpolation and extrapolation. 

\begin{table}[!ht]
\vspace{0.1cm}
\tabcolsep=5pt\fontsize{10}{16}\selectfont
\centering
\captionsetup{font=small}
\caption{The average and standard errors of the root mean square geodesic errors of WGPR, iGPR and MGPR for flight trajectory data.}
\label{table5}
\begin{tabular}{cccccc ccccc c}
\hline
 & \multirow{1}{*}{Scheme}   & $N$  & \multicolumn{1}{c}{iGPR} & \multicolumn{1}{c}{WGPR} & \multicolumn{1}{c}{MGPR} \\\hline
& Random & 30    & 0.0237(0.0382)    &0.0539(0.0571)      &1.0293(0.5219)     \\
&  & 50          & 0.0187(0.0301)    &0.0364(0.0478)      &0.8749(0.4392)     \\
&  & 100         & 0.0105(0.0226)    &0.0266(0.0302)      &0.8223(0.3927)     \\
&Sort  & 30      & 0.0583(0.0334)    & 0.0877(0.0729)  & 1.2245(0.5649)      \\
&  & 50          & 0.0473(0.0267)    & 0.0702(0.0688)  & 1.3190(0.5753)      \\
&  & 100         & 0.0211(0.0209)    & 0.0495(0.0555)  & 1.0029(0.4966)      \\  \hline
\end{tabular}
\vspace{-0.5cm}
\end{table}

In addition, we plot the estimated flight trajectory on 10 August 2024, using two methods, WGPR and iGPR, as shown in Figure of Section 10 of the Supplementary Materials. Under the Sort sampling scheme, the total number of observation points is $N=40$, with the last three points designated as test data. Compared to WGPR, the estimated trajectory from iGPR is closer to the true one, indicating that our method, iGPR, is more effective at modeling manifold data because of its intrinsic property. 

\subsection{Diffusion tensor imaging data}
\begin{figure}[!ht]
\vspace{-0.7cm}
\centering
\subfigure[True DTI dataset]{
\begin{minipage}[b]{0.45\linewidth}\label{fig103a}
\centering
\includegraphics[scale=0.25]{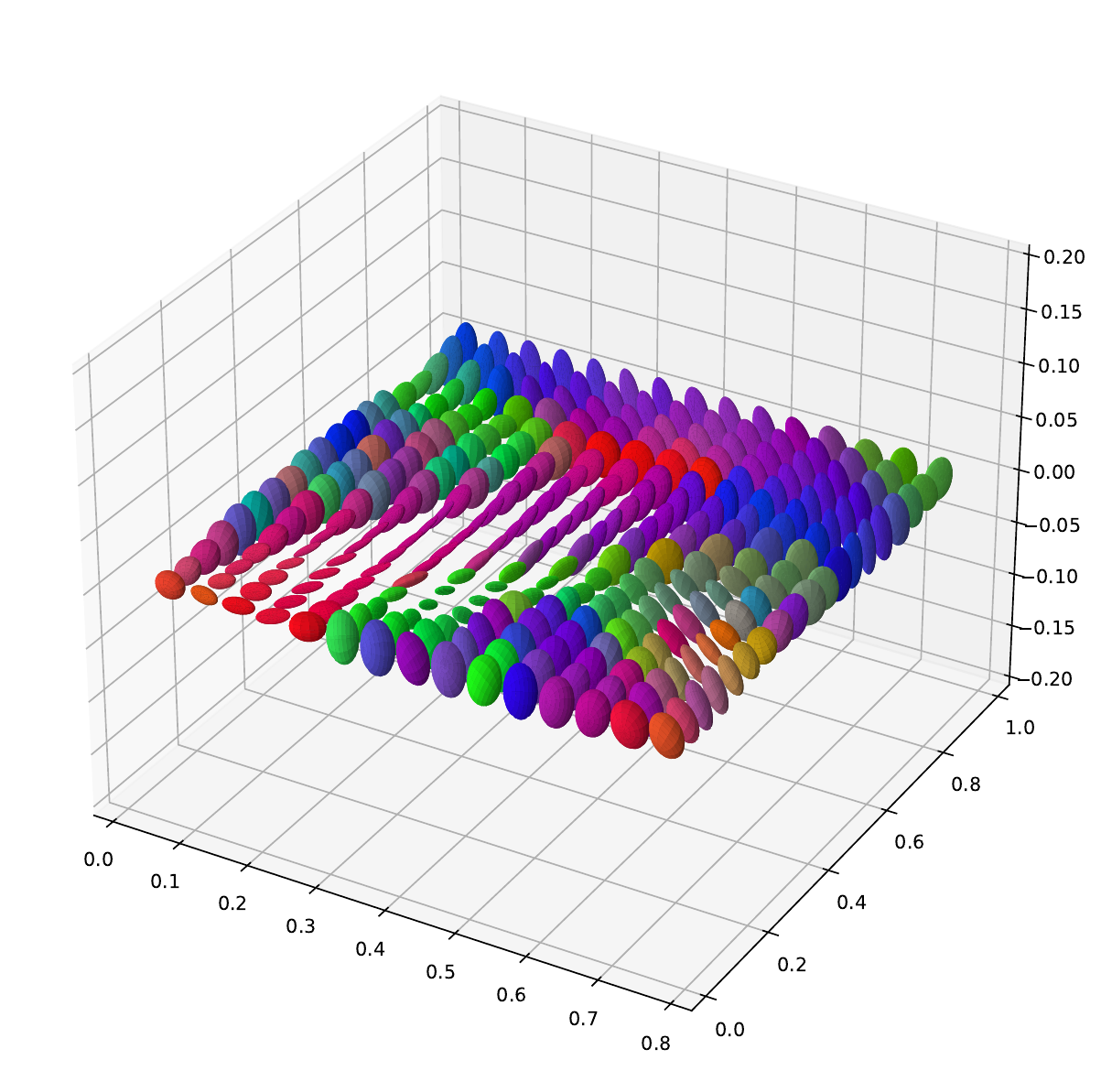}
\end{minipage}
}
\subfigure[Reconstruction of DTI dataset]{
\begin{minipage}[b]{0.45\linewidth}\label{fig103b}
\centering
\includegraphics[scale=0.25]{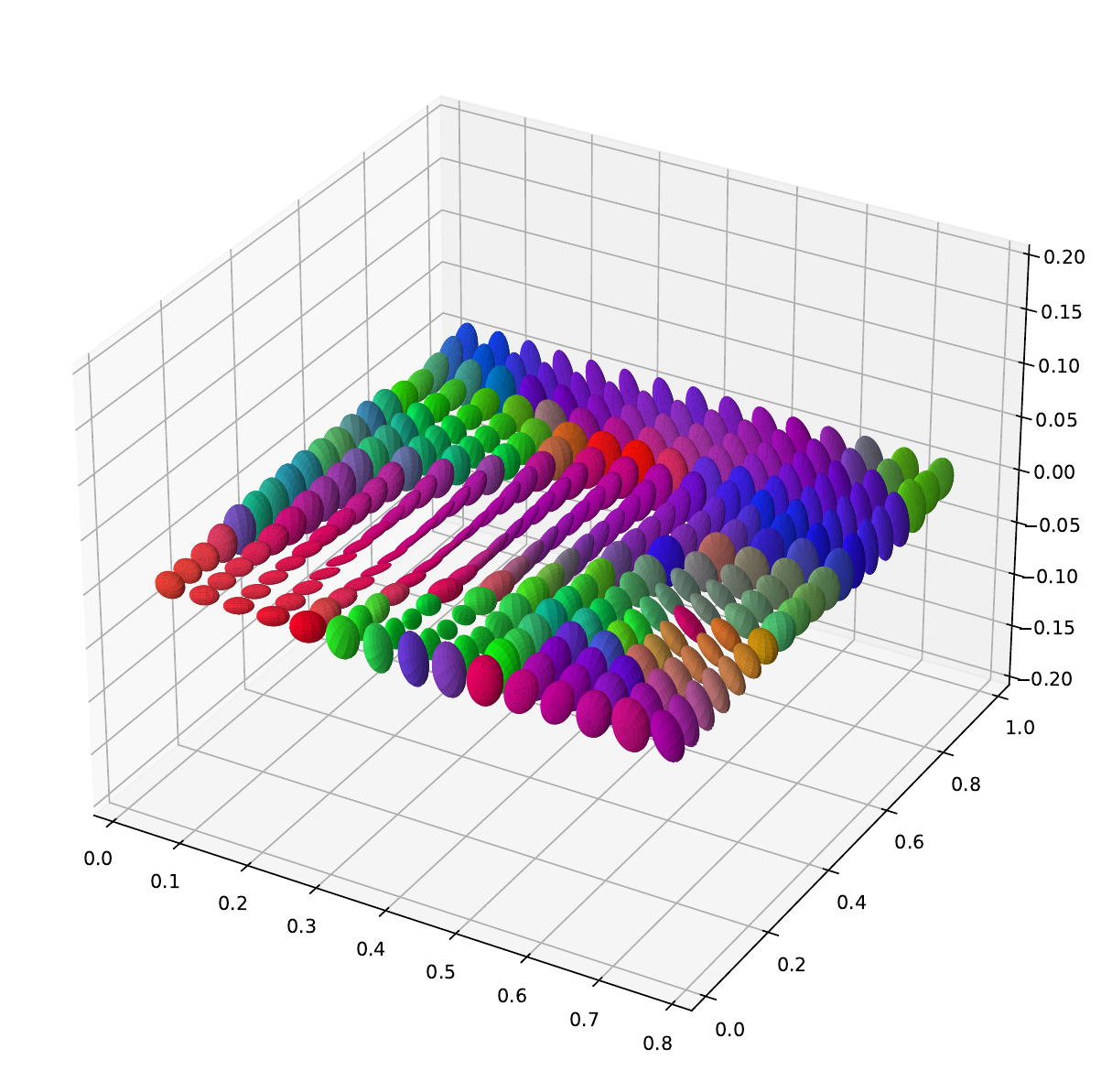}
\end{minipage}
}
\caption{The three-dimensional ellipsoid visualization of the sampled DTI dataset.}
\vspace{-0.5cm}
\end{figure}

We consider another real data set for a patch of estimated voxel-wise DTI tensors from a coronal slice of an HCP subject \cite{glasser2013minimal}. The responses reside on the manifold $\mathcal{S}_{++}^6$ and have a non-positive curvature. This data structure can be fitted using the iGPR method. The predictors are two-dimensional index variables, which take values in $\mathbb{Z}^{2}_{+}$. The data used in this study can be downloaded from (\url{https://drive.google.com/uc?export=download&id=1psVU_kw86gqYLD9TlyLGvFba61dlWayB}). A visualization of the response for DTI data is shown in Figure \ref{fig103a}. The ellipsoid represents the diffusion characteristics of the DTI tensors, with its shape, size, and orientation determined by the eigenvalues and eigenvectors of the tensors. Additionally, the color represents the relative magnitude of the maximum singular value.

\begin{table}[!ht]
\centering
\tabcolsep=5pt\fontsize{10}{16}\selectfont
\captionsetup{font=small}
\caption{The root mean square geodesic errors of WGPR and iGPR for Diffusion tensor imaging data.}
\label{table10}
\begin{tabular}{cccccccccc}
\hline
  & \multirow{1}{*}{Sampling ratio} & \multicolumn{1}{c}{iGPR}  & \multicolumn{1}{c}{WGPR}  & \multicolumn{1}{c}{MGPR}\\\hline
                           & 25\%                & 0.8650 &1.859   &5.3648\\
                           & 50\%              & 0.5595 &1.433   &4.7528\\
                           & 75\%               & 0.2112 &1.211   &5.0920  \\\hline
\end{tabular}
\end{table}

We selected 25\%, 50\%, and 75\% of the total samples, evenly spaced, as train data, while the remaining samples were used as test data. The primary objective of this study is to reconstruct a patch of estimated voxel-wise DTI tensors from a coronal slice of a subject. To support this reconstruction process, we employed geodesic regression as the prior BPF and used a diagonal Radial Basis Function (RBF) kernel as the TSCF in iGPR. Table \ref{table10} presents the root mean square geodesic errors from iGPR, WGPR and MGPR for DTI data.  We see that iGPR has considerably smaller estimation errors compared to WGPR and MGPR, highlighting its superior performance in reconstruction accuracy. Moreover, Figure \ref{fig103b} shows the estimated three-dimensional visualization of the ellipsoid using only 25\% of the samples, demonstrating that iGPR successfully reconstructs the tensor field.

\section{Conclusions}\label{s3}
Recognizing that directly defining a covariance function in the ambient space may undermine the inherent geometric properties of a manifold, this paper employs the parallel transport operator to create a new structure of the covariance function with desirable statistical properties. We then develop an intrinsic Gaussian process regression model to estimate the regression function and depict the uncertainty. For applications, computation of covariance function of a intrinsic Gaussian process usually needs an orthogonal coordinate frame for representation. We show that the posterior distribution of the regression function is independent of the chosen orthogonal coordinate frame. Furthermore, we establish the statistical consistencies of the posterior distribution estimator, including the information consistency and posterior consistency.  Numerical studies show that iGPR is effective in analyzing data with Riemannian manifolds, providing meaningful uncertainty estimates.

Our method accommodates data with response in the manifold space and predictors in Euclidean space. The exploration of more complex spaces, such as metric spaces, is relatively limited in the existing literature. We anticipate that our proposed method would shed light on the analysis of data in complex spaces. In future work, we plan to extend the idea to data where the response is in manifold space and predictors are in non-Euclidean space, developing additive models based on iGPR. Although the posterior distribution in this paper has explicit expression, it is no longer a iGD, because the posterior mean ${\boldsymbol{\beta}}_{t^{*}}^*$ might not be zero. It raises a question of whether when performing GP regression on manifolds, we should consider other extensions that might be more appropriate than the extension. Based on non-Euclidean structure of manifolds, constructing a normal distribution directly on a Riemannian manifold without using the Riemannian exponential map is significantly challenging. However, this area is worth further study.

\vspace{-0.3cm}
\section*{Acknowledgements}
This research is supported in part by funds of National Key R\&D Program of China (2023YFA1011400), National Natural Science Foundation of China (No. 12371277, 12231017, 12271239, 12201601), and Shenzhen Fundamental Research Program JCYJ20220818100602005 (No. 20220111).

\bibliographystyle{abbrvnat}
\bibliography{Intrinsic_GP}

\end{document}